\documentclass{vgtc}                          

\ifpdf
  \pdfoutput=1\relax                   
  \usepackage{graphicx}                
  \DeclareGraphicsExtensions{.pdf,.png,.jpg,.jpeg} 
\else
  \usepackage{graphicx}                
  \DeclareGraphicsExtensions{.eps}     
\fi%

\graphicspath{{figures/}{pictures/}{images/}{./}} 

\usepackage{microtype}                 
\PassOptionsToPackage{warn}{textcomp}  
\usepackage{textcomp}                  
\usepackage{mathptmx}                  
\usepackage{times}                     
\usepackage{cite}                      
\usepackage{tabu}                      
\usepackage{booktabs}                  
\usepackage{flushend}  

\onlineid{7354}

\vgtccategory{Research}

\vgtcinsertpkg

\title{AI Total: Analyzing Security ML Models with Imperfect Data in Production}

\author{Awalin Sopan\thanks{e-mail: awalin.sopan@sophos.com}\\ %
        \scriptsize Sophos %
\and Konstantin Berlin\thanks{e-mail: konstantin.berlin@sophos.com }\\ %
     \scriptsize  Sophos %
     }

\teaser{
  \centering
  \includegraphics[width=\linewidth]{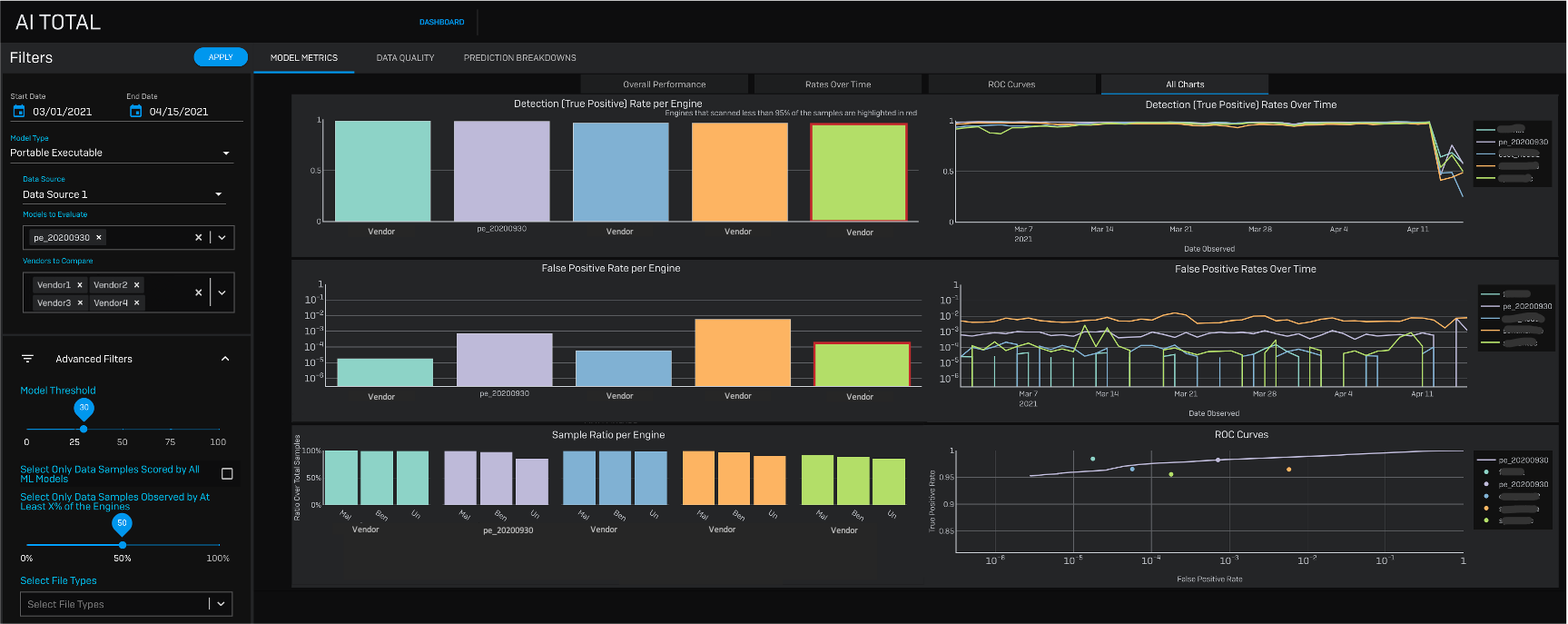}
  \caption{The landing page of the dashboard, showing the Filter panel to the left, and the Visualization panel with three tabs on the right, with the Model Metrics tab selected. The first two charts are Detection (True Positive) Rates and False Positive Rates of our selected models and other anonymized detection engines for detecting malicious portable executable (PE) files. Then we show the Ratio of Samples scanned per engines, followed by TPR, FPR and the finally the ROC curve.}
  \label{fig:model-metric}
}

\abstract{Development of new machine learning models is typically done on manually curated data sets, making them unsuitable for evaluating the models’ performance during operations, where the evaluation needs to be performed automatically on incoming streams of new data. Unfortunately, pure reliance on a fully automatic pipeline for monitoring model performance makes it difficult to understand if any observed performance issues are due to model performance, pipeline issues, emerging data distribution biases, or some combination of the above. With this in mind, we developed a web-based visualization system that allows the users to quickly gather headline performance numbers while maintaining confidence that the underlying data pipeline is functioning properly. It also enables the users to immediately observe the root cause of an issue when something goes wrong. We introduce a novel way to analyze performance under data issues using a data coverage equalizer. We describe the various modifications and additional plots, filters, and drill-downs that we added on top of the standard evaluation metrics typically tracked in machine learning (ML) applications, and walk through some real world examples that proved valuable for introspecting our models.%
} 

\begin{document}


\firstsection{Introduction}

\maketitle

Using machine learning (ML) for cyber security applications has got a lot of traction at present \cite{blowers2014machine, ledoux2015malware, kyadige2020learning, saxe2015deep,anderson2016deepdga}, from detecting malicious files, URLs, domains, emails etc., to aiding analysts in Security Operation Centers (SOC) \cite{sopan2018building, angelini2017goods, DELORENZO2020102423}. Organizations deploy ML models in their systems to detect and block potential malwares, and cyber-security analysts triage security threats with the help of ML. Badly functioning models can quickly result in too many false detections – to the point where the model might have to be deactivated. Alternatively, a bad model may fail to detect a serious malicious attack. In typical ML deployments, the data distribution rarely changes, and the model is deployed in a ``set and forget" environment. In a controlled model development phase, data scientists develop models on carefully curated static data sets. But in an ever-changing cyber-landscape with the dynamic data \cite{wagner2015survey} with its mission-critical applications, a model needs to be continuously monitored, and updated whenever required to remain relevant.

Once a security model is deployed, and starts being evaluated on incoming data streams, the evaluation can run into unforeseen scenarios with engineering issues occurring at various stages of production, may it be during data ingestion, data aggregation, or data scoring. Properly evaluating the models becomes challenging when we do not understand what is impacting the distribution and labeling of evaluation data. Large organizations often have separate data infrastructure teams, data science teams, and malware researcher teams, each focusing on different aspect of the products. But this can cause further confusion among teams when rapid response is needed instead. Considering all these situations, we built a real-time visualization dashboard, AI Total (\autoref{fig:model-metric}) that helps our security data science team to fulfil these goals:
\begin{itemize}
\item Monitor the deployed models’ regular performance, observe time trends, and detect anomalies
\item Detect issue with the evaluation data, models, or labeling
\item Investigate the reasons behind the issues
\end{itemize}

In this paper, we present this visualization system, and discuss how it fulfilled those goals. We gathered the requirements from our agile development process and observed the utility of this tool by its direct usage over time. We share some cases illustrating the practical value of this application and the lessons learned from its usage. 

\section{Related Work}
A good amount of work features visualizations that focus on model training and parameter tuning \cite{wongsuphasawat2017visualizing, chatzimparmpas2021featureenvi}. There are important works on direct error examination \cite{amershi2015modeltracker}, data validation \cite{breck2019data, sambasivan2021everyone} and label validation \cite{cordeiro2020survey} but they all focus on the development phase alone. Maintaining a large-scale machine learning system with industry scale data comes with its own challenges \cite{kahng2017cti, bosch2021engineering, amershi2019software}. Monitoring such systems involves data, prediction, and system monitoring \cite{breck2017ml}. Uber’s Michelangelo platform \cite{hermann2017meet} enables teams to monitor model performance but does not provide a way to correlate model performance issues to the data issues. Aporia \cite{aporia} helps with detecting data integrity and data drift but cannot drill down to specific part of models, and its many charts presented at once can cause information overload. Our work focuses on what happens after a model is deployed in production and we designed the system specifically for security ML models while keeping the visualizations simple and easy. It helps correlate model issues with data issues and shows the breakdown of the model performance based on cyber-domain specific categories.

\section{Application Background}
Our team develops machine learning models that predict maliciousness of different types of artifacts, such as files, URLs, emails, etc. Deploying the models in production requires our team to maintain situational awareness in order to identify systematic issues with deployed models, measuring when the model should be retrained, and identifying issues with incoming data that is used for evaluation and training of future models. The dashboard users are our data scientists, data engineers and business stakeholders. 

\textbf{Data Labeling:} To evaluate the model performance, we label the data based on a complex set of rules that take third party vendor results and internal data into consideration. We combine information about the artifacts from multiple sources. Some internal information used in the labeling function are the number of times an artifact has been seen across customer base, age of the artifact, sandbox reports that observed the artifact, and matches on handwritten signatures. For external vendor results we take a weighted majority voting on the verdicts on the artifacts from various industry vendors. The labels for artifacts changes over time as we gain more context regarding them, and the model performance also changes. 

\textbf{Data Pipeline:} AI Total is built on the backbone of a highly scalable data-aggregation and data-correlation process. To evaluate the model performance, various internal and external telemetry streams are collected, aggregated and correlated daily. The model results are combined with internal telemetry and reputation services to provide new evaluation data and updated labels. The reputation system, external vendors, and our models will be referred to as ‘engines’.

\section{Web Application Interface}
The first goal behind AI Total was to provide the high-level users with situational awareness of the models' performance. A gradual decay in performance indicates a concept drift and a sudden drop signals a sudden change in incoming data. Upon observing such issues, users may consider retraining the model. But the problem can rather lie in the data and hence the performance result itself might be wrong. Moreover, when a system is missing labels or model-scores for a large portion of the data set, its performance metrics are not reliable either. So, advanced users want to view the data-health and the model performance in a more detailed breakdown, to correlate model issues with data issues, and to have the ability to filter and sub-select data of interest without writing complex queries. Since the system would be used by various levels of users with tasks involving simple monitoring to complex exploration, we presented the filtering options and the charts in a progressive manner in the user interface (UI). The two main UI components are: 1. The Filter panel and 2. The Visualization panel.
\subsection{Filter Panel}
The Filter panel (\autoref{fig:model-metric}, left) enables users to generate custom queries for the dashboard charts. This panel is always visible, so users can know what selection of data they are analyzing. Users can select the \textbf{Source} of data feed and the \textbf{Time frame} for which they want to view the data, \textbf {Model types} that predict on different types of artifacts, namely, Portable Executable (PE), Microsoft Office, PDF, URL, etc. After that, different \textbf{Versions of the models} and \textbf{Vendors} they want to compare with. By default, the users see the latest deployed model evaluated on the last two weeks of data.

The Advanced Filter component (collapsed by default) is used by advanced users to filter the data set in more complex ways. Here, users can adjust the \textbf{Model threshold} for prediction using a slider, and immediately see how the True Positive Rate (TPR), False Positive Rate (FPR), and ROC (receiver operating characteristic) curve change and choose a new threshold to get the best result. They can also choose only certain \textbf{Types of files} from their analysis. 

One of the major issues we faced when comparing different engines is that engines are not always scored on the same artifacts. This could heavily bias the analysis depending on the reason for why some engines are not scoring. To compare engines' performances, we need to measure them against the same data. However, it is possible for one of the selected engines to barely have any scores, thus while the distribution is equalized it is hardly an accurate representation of the true desired data distribution. To balance those two factors, we have two special controls to \textbf{equalize the data coverage}: a check box to \textit{select only the data scored by our model}, and render the charts for that subset of the data, and a slider to configure \textit{the \% of engines that observed the data}: such that at x\% mark, it will render the metrics on only the artifacts scored by x\% of the engine. This allows us to focus only on the subset of data that is relevant for evaluating our models. In \autoref{fig:model-metric}, we have set this at the 50\% mark, so the data used for the analysis has been seen by at least half the engines. To evaluate the engines against the same data, we can equalize the data coverage by selecting only data samples observed by 100\% of the engines. This way the comparison will be more accurate, but this may result in significant data shrinkage with less reliable results. With no constraint (slider at 0\%) the charts will be shown for all incoming data irrespective to the engines' coverage.

\subsection{Visualization Panel}
The Visualization panel consists of three tabs, each focusing on specific goals:

\textbf{Model Metrics:} shows an overview of the performance of the models and vendors. Used by CTO, scientists, engineers, etc.

\textbf{Data Quality:} presents data volume and rate to help monitor data issues that affects the models. Used by data scientists, engineers, DevOps, and infrastructure engineers.

\textbf{Prediction Breakdown:} shows a breakdown of the model’s performance in several relevant categories, corresponding data volume, and a few aggregate-statistics to help investigate issues. Used by data scientists.

We followed the overview, zoom and filter, details on demand principle \cite{ben} here. To view the details, users can click a chart element and view a table view of the data generating that element and can download the data in spreadsheet format for further analysis. 

\subsubsection{Model Metrics tab}
The models’ performance is our core focus, hence the first view is the Model Metrics tab (see \autoref{fig:model-metric}) showing the selected engines' overall TPR Detection and FPR, Scanned Data ratio plot, TPR and FPR over time, and the ROC curve of the models. The engines are mapped in a categorical color scheme. Detecting malwares correctly and reducing false positives are key performance indicators for security industry, so we broke down the model performance into these two metrics upfront. 

The engines failing to scan a lot of incoming data indicates a potential issue; to highlight that case, a red border is drawn around the bars for engines that scanned less than a certain \% of the data, indicating that our scoring mechanism may have broken at some point, and we cannot rely on performances of such engines with a low data coverage. To check it further, the Sample Ratio per Engine shows how much of the ingested data each engine has scanned; with bar charts showing the percentage of data scanned, grouped by their labels: malicious or benign, or unlabeled. Here, users observe what \% of data each engine is scanning. It is concerning when a model has a low value for the sample ratio plot compared to the other vendors. Then the users can adjust the data coverage using the slider in the Advanced filter panel; for example, they can increase the coverage requirement such that the plots will be calculated only for the data the model has scored, or at least a large portion of the engines have scored. With this change, the Sample Ratio chart and the corresponding FP and TP charts will change as well. If all the engines show a low percentage for data they managed to scan and score, then the system does not have verdicts from the models or the vendors. In this situation, the labels cannot be trusted either since the labeling depends on the majority voting of external vendors. 

While the overall performance over a selected time range provides an aggregate view of the performance, it is not detailed enough to help identify temporary issues or inflection points. Our next charts, TPR and FPR over time, are for this purpose of comparing a model's current performance with its historical behavior. A gradual decay of a model’s performance can be due to a concept drift as the feature distribution of incoming data may change from the training data, signaling it is time to retrain the model with new data. In \autoref{fig:model-metric}, we see that, our selected model, PE\_20200930 is performing better than 3 of the 4 selected vendors in detecting malicious artifacts (TP chart), but also generating a lot of false positives (FP charts). Its performance suddenly dropped at the end (TPR over Time per Engine chart). Users can increase the prediction threshold to check if that reduces the FPs. Sometime a model generates a lot of FPs when it detects some emerging threat that the current malware detection industry is missing. To investigate this further, users use the Details view to get more context of these FP artifacts, download the metadata, and send to the threat research team for cross-check with their intel. 

\subsubsection{Data Quality tab}
\begin{figure}[tb]
 \centering 
 \includegraphics[width=\columnwidth]{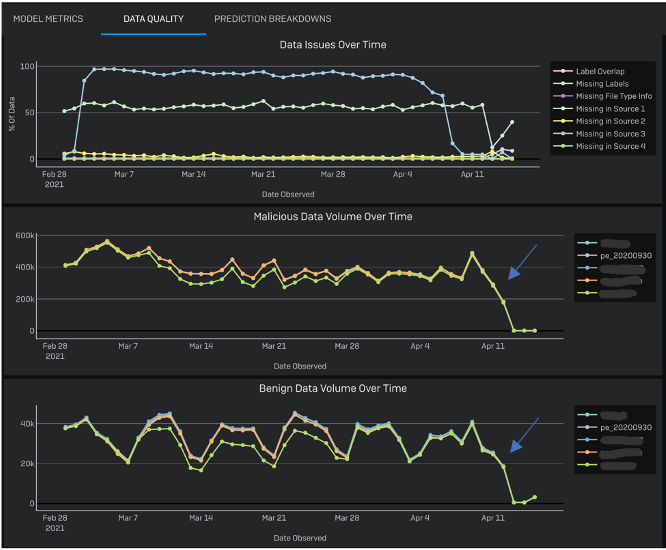}
 \caption{The Data Quality tab is in focus in the Visualization panel. A sudden drop of data volume after the first week of April in Malicious and Benign Data Volume over Time charts indicate a potential problem in the data feed. Vendor names are hidden. Data shown here is fabricated, for illustration purpose only. }
 \label{fig:data-quality}
\end{figure}

In production, often the model’s first point of failure can be in the data pipeline itself. To monitor that, our next view focuses on Data Quality (see \autoref{fig:data-quality}) showing various data related metrics. The top chart, called Data Issue over Time, shows time series of various data quality metric. The high-level goals of this view are to monitor data consistency and to detect anomaly in input sources and in data labels. Here we show the percentage of data pertaining to: data missing from input sources, data with no labels, and data with missing file type information. Seeing high volume of missing label indicates our labeling system needs attention and the model performance is not reliable since we do not have enough labeled data. Seeing data missing from input source feeds points us to check the data provider system. Other charts show the volume of malicious, benign, and unlabeled data that are scanned by the engines per day. A sudden drop in incoming data volume may signal that our telemetry ingestion system is down (\autoref{fig:data-quality}). A sudden spike of incoming data volume hints a change in data distribution, which may also influence the model’s performance. 

\begin{figure}[tb]
 \centering 
 \includegraphics[width=\columnwidth]{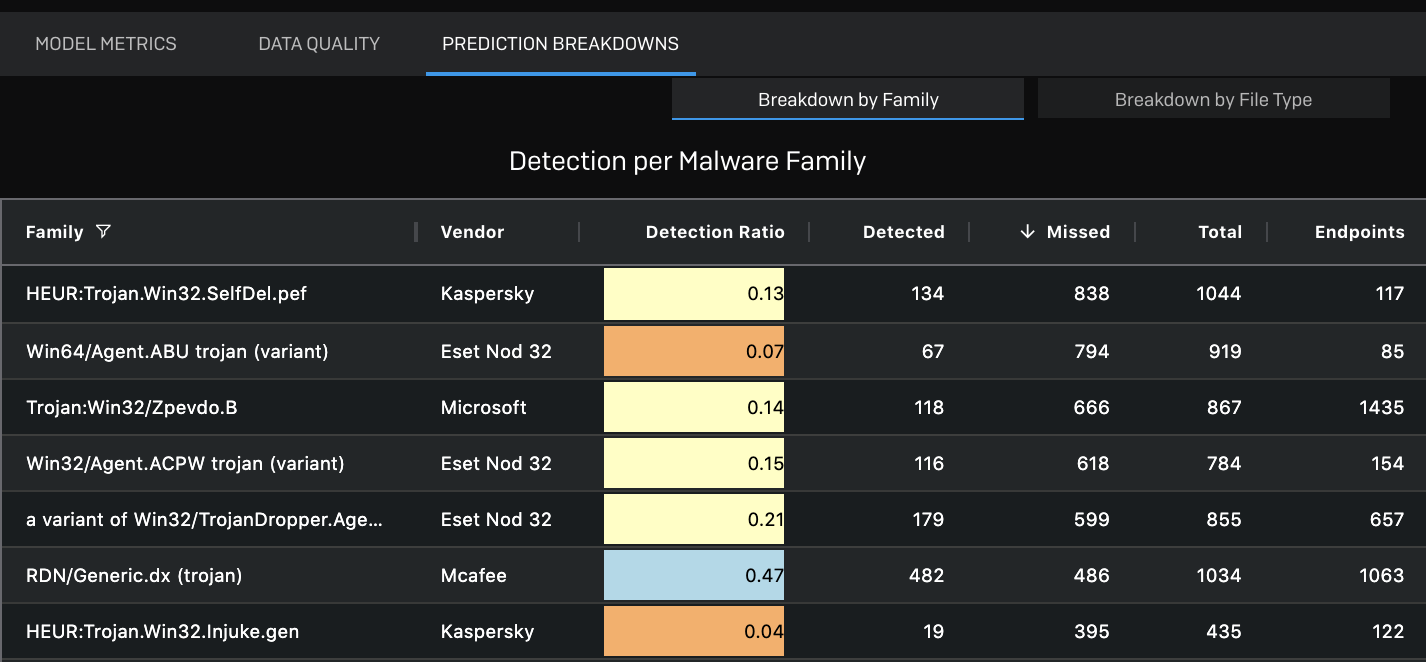}
 \caption{The Prediction Breakdown tab in the Visualization panel. The table in focus is the PE malware artifacts grouped by malware family names, here filtered to show only Trojan files, sorted by detection missed by our model. The columns are: the Vendors providing the name, the Detection Ratio of that family by our model, the amount of Detected (TP), Missed, Total samples, and Endpoints (machines affected in customer environments).}
 \label{fig:pred-break}
\end{figure}

\subsubsection{Prediction Breakdown tab}
Similar to the FP detailed window, sometimes it is helpful to provide views based on more fine grain grouping of data. Such a view is particularly important for security ML researchers. In the Prediction Breakdown tab we have two tables (\autoref{fig:pred-break}), one showing data grouped by malware family, and another grouped by file types. The tabular views support multivariate data with several thousand rows. An average performing model can be stronger for a certain subset of the data set, and even a strongly performing model can have poor accuracy in some subset of the data set. While the Model Metrics view shows the model's overall TP and FP rates, here we show the model's accuracy across some domain-specific categories. To highlight low performing zones, we color coded the detection ratio cell using a divergent scheme: detection rate high to low is blue to orange. If the model is failing to detect a specific type of malware family the issue can be either some missing feature or even parsing content of that specific family. Users can also filter the table based on column values; for example, they can select only malware families that contain \textit{‘emotet’} in their names. This way, if there is a sudden outbreak of a new malware family, they can filter to only the files belonging to that family and can quickly check if the model is able to detect that. Users can view the Details of each row in the table by clicking the row and download sample data.

\begin{figure}[tb]
 \centering 
 \includegraphics[width=\columnwidth]{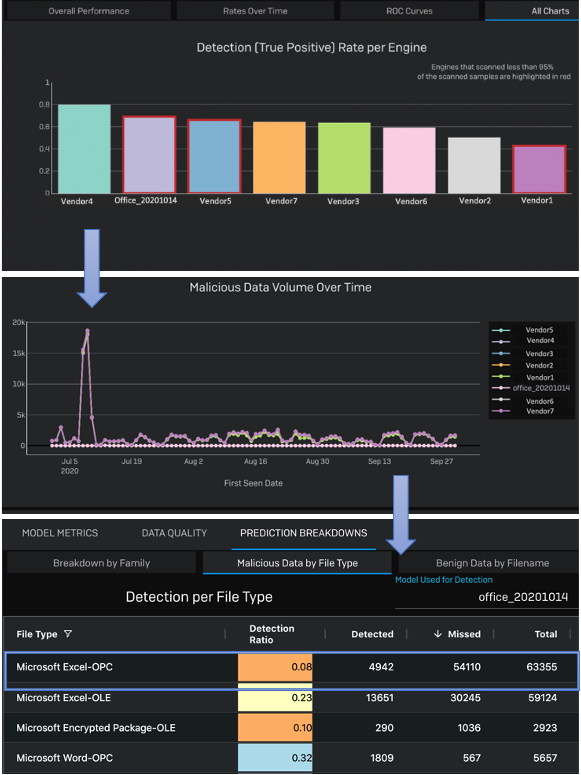}
 \caption{Top: The Office model has a very low TP and red warning border indicating low data coverage. Middle: The Data Quality tab shows a spike of incoming data in July. Bottom: The Detection per File type table shows most of those files are Microsoft Excel files (Total column), and most of the files our models missed to detect are also of the same type (Missed column), and the Detection ratio (ratio of malicious files detected by our model and actual number of malicious files of that type) is also low.}
 \label{fig:office}
\end{figure}
\section{Use Cases}
In this section we illustrate two specific cases where the tool was useful to investigate some problems.

\paragraph{Troubleshooting Office Model Low Performance:} Our data scientists noticed a sudden performance drop of our models (in the Model Metrics view) that detect malicious Microsoft Office files and a red border around the TP and FP bars around our model indicated that the model did not manage to scan all the data. Switching to the Data Quality view, they saw an unexpected spike in number of incoming files and most of these files are Excel files. This new incoming data distribution is different from the distribution in the regular training data, affecting the current model’s performance. Then they used the Prediction Breakdown view per file type, sorted the table by the descending number of missed detections, and noticed that the model’s performance is low for Excel-OPC documents (\autoref{fig:office}, top). The color-coded cells in the Detection ratio column helped highlight the issue very quickly. All these various indicators allowed our data scientists put things together and they decided to re-evaluate the model without the Excel files. After excluding Excel documents from analysis (using the Filter by File Type), they saw that the model’s performance was on track if they ignore the anomalous surge in malformed MS-Excel files over the past month that our model failed to scan. This granular comparison helped them to make an appropriate decision. 

\paragraph{Fix parsing of PDF Files:} One of our models detects malicious PDF files. A data scientist looked at the model’s prediction breakdown by Malware Family and noticed that the model had a lower detection rate for the files from ‘RDN/Generic.cf’ family. They opened the Details view to check the sample files and realized that the system was unpacking Gzip files with PDFs inside them, however the model then tried to analyze the Gzip as a PDF without unpacking and thus made wrong prediction on them. After spotting this problem using the tool, the team fixed the parsing system.

\section{Lesson Learned}
We improved the UI gradually according to user feedback. We noticed that the high-level users do not usually interact with the filters rather observe the default Model Metric view for reporting purpose, so we needed to put enough information in the UI such that even as static views the charts are fully informative with proper labels. When the charts do not show any data, the users needed to know if it is a system failure, or the result of their query just happens to be empty. So, we improved the UI feedback with better error messages in case of an empty query result, and added query progress feedback in case of a slow query. The strengths of this system can be summarized as follows:
\par 1. We presented the views in the UI tabs progressively in order from ‘simple and most frequently used’ to ‘complex and used when needed for further investigation’, so the first tab is the Model Metrics: used by most of the users, from data scientists to high-level users like upper management on a weekly basis; then Data Quality: used by data scientists and engineers knowledgeable in the system architecture; and finally the Prediction Breakdown: used by data scientists to debug the models. The Details on Demand feature was requested by threat researchers; while it is rarely used, it is deemed very useful whenever it is.
\par 2.  The comparison among the engines is not perfect unless they are measured against the same data. But we cannot trust the performance of the engines if only a small portion of the data is seen by them. Hence, we added the red border for highlighting low data coverage and the Sample Ratio chart to view and compare the data coverage. To evaluate the models within that imperfection we added the Data coverage filter. This novel filtering and chart combination helped the team evaluate the results from different angles.
\par 3. Identifying the root-cause of an issue also helps identify which teams to reach out for support. For example, the infrastructure team may help fix the pipeline or back fill any missing or corrupted information. So, we familiarized the data infrastructure and malware researcher team-members with the tool and included their feedback as well. We agreed on a common vocabulary to use for the UI labels for a better cross-team collaboration.
\par 4. We added multiple simple views rather than one complex view to support data scientists' workflow while keeping it simple for high-level users. We focused on finding trends and anomalies in data feeds relevant to the models. A combination of several charts enabled the team to ask questions, verify their hypotheses and generate insights. 

\section{Conclusion}
In this paper, we described how the power of visualization fulfilled the operational needs of our industry research team to detect and resolve the types of issues we frequently see in productionized operational security models. As the project matures, we hope to perform a more in-depth evaluation of the tool and a design study. For now, we shared lessons learned to help other security ML teams designing their monitoring infrastructure in a systematic way. 

\acknowledgments{
The authors wish to thank Richard Harang, Ajay Lakshminarayanarao and Alex Long for their contribution to the project.}

\bibliographystyle{abbrv-doi}
\bibliography{aitotal}
\end{document}